\documentclass{article}
\usepackage{geometry} 

\usepackage{xspace}
 
\usepackage[utf8]{inputenc} 
\usepackage[T1]{fontenc}    
\usepackage{hyperref}       
\usepackage{url}           
\usepackage{booktabs}
\usepackage{amsfonts}
\usepackage{nicefrac} 

\usepackage[title]{appendix}
\usepackage{times}
\usepackage{graphicx}
\usepackage{amsmath}
\usepackage[usenames,dvipsnames]{color}
\usepackage{multirow}
\usepackage{authblk}
\usepackage{caption} 
\usepackage{subcaption}

\newcommand{\jack}[1]{}
\newcommand{\rashmi}[1]{}
\newcommand{\shivaram}[1]{}

\newcommand{\etal}{et al.\xspace}
\newcommand{\encodef}{$\mathcal{E}$\xspace}%encoding function
\newcommand{\decodef}{$\mathcal{D}$\xspace}%decoding function
\newcommand{\func}{\mathcal{F}\xspace}
\newcommand{\done}{X_1\xspace}
\newcommand{\dtwo}{X_2\xspace}
\newcommand{\dk}{X_k\xspace}
\newcommand{\mlpcoder}{MLPEncoder\xspace}
\newcommand{\convcoder}{ConvEncoder\xspace}
\newcommand{\FC}{FC:\xspace}
\newcommand{\Conv}{Conv:\xspace}
\newcommand{\dilation}{dilation:\xspace}
\newcommand{\channelmodel}{base model\xspace}
\newcommand{\Channelmodel}{Base model\xspace}
\newcommand{\ChannelModel}{Base Model\xspace} % CHANGE THIS IN TANDEM WITH ABOVE
\newcommand{\mnist}{MNIST\xspace}
\newcommand{\fashion}{Fashion-MNIST\xspace} % CHANGE THIS IN TANDEM WITH BELOW

\newcommand{\cifar}{CIFAR-10\xspace}
\newcommand{\ecacc}{recovery-accuracy\xspace} % CHANGE THIS IN TANDEM WITH THE 2 BELOW
\newcommand{\ECacc}{Recovery-accuracy\xspace}

\newcommand{\ecaccs}{recovery-accuracies\xspace}
\newcommand{\trueacc}{overall-accuracy\xspace} % CHANGE THIS IN TANDEM WITH THE 2 BELOW
\newcommand{\Trueacc}{Overall-accuracy\xspace}

\newcommand{\classificP}{\mathcal{C}\xspace}

\newcommand{\channelMLP}{Base-MLP\xspace}
\newcommand{\channelResnet}{ResNet-18\xspace}

\newcommand{\lossShort}{Training Loss Func.\xspace}
\newcommand{\lossMSE}{MSE-Base\xspace}
\newcommand{\lossKL}{KL-Base\xspace}
\newcommand{\lossTrue}{XENT-Label\xspace}

\newcommand{\outputReconstruct}{\widehat{\func(X)}}
\newcommand{\outputReconstructOne}{\widehat{\func(\done)}}
\newcommand{\outputReconstructTwo}{\widehat{\func(\dtwo)}}
\newcommand{\outputReconstructK}{\widehat{\func(\dk)}}
\newcommand{\outputChannel}{\func(X)}

\newcommand{\linearCodedComputeRefs}{\cite{kangwook-og,dutta2016short,li2016unified,yu2017polynomial,wang2018sparse,dutta2017coded,reisizadeh2017coded,mallick2018rateless}\xspace}

\title{Learning a Code: Machine Learning for Approximate \\ Non-Linear Coded Computation} 

\date{}
\author[1]{Jack Kosaian}
\author[1]{K.V. Rashmi}
\author[2]{Shivaram Venkataraman}
\affil[1]{Computer Science Department, Carnegie Mellon University}
\affil[2]{Microsoft Research}

\affil[1]{\{jkosaian, rvinayak\}@cs.cmu.edu}
\affil[2]{shivaram.venkataraman@microsoft.com}

\begin{document}

\maketitle

\begin{abstract}
Machine learning algorithms are typically run on large scale, distributed compute infrastructure that routinely face a number of unavailabilities such as failures and temporary slowdowns. Adding redundant computations using coding-theoretic tools called ``codes'' is an emerging technique to alleviate the adverse effects of such unavailabilities. A code consists of an encoding function that proactively introduces redundant computation and a decoding function that reconstructs unavailable outputs using the available ones. Past work focuses on using codes to provide resilience for linear computations and specific iterative optimization algorithms. However, computations performed for a variety of applications including inference on state-of-the-art machine learning algorithms, such as neural networks, typically fall outside this realm. In this paper, we propose taking a \textit{learning-based} approach to designing codes that can handle non-linear computations. We present carefully designed neural network architectures and a training methodology for learning encoding and decoding functions that produce \textit{approximate} reconstructions of unavailable computation results. We present extensive experimental results demonstrating the effectiveness of the proposed approach: we show that the our learned codes can accurately reconstruct $64 - 98\%$ of the unavailable predictions from neural-network based image classifiers (multi-layer perceptron and ResNet-18) on the MNIST, Fashion-MNIST, and CIFAR-10 datasets. To the best of our knowledge, this work proposes the first learning-based approach for designing codes, and also presents the first coding-theoretic solution that can provide resilience for any non-linear (differentiable) computation. Our results show that learning can be an effective technique for designing codes, and that learned codes are a highly promising approach for bringing the benefits of coding to non-linear computations.

\end{abstract}

\section{Introduction}

Machine learning serves as the backbone for a wide variety of cognitive tasks such as image classification, object recognition, and natural language processing. Today, applications can leverage state-of-the-art machine learning models by using cloud services that offer machine learning as a service~\cite{azure-ml, google-ai, aws-ml}. To handle large traffic, such service providers typically use a distributed setup with a large number of interconnected servers (compute nodes).  It is well-known that such a distributed compute infrastructure faces a number of unavailability events~\cite{jeff-dean-failures,rashmi2014hitchhiker,asterisxoring}. First, these clusters are typically built out of commodity components making failures the norm rather than the exception. Second, various factors including load imbalance and resource contention cause transient slowdowns. (Servers facing such temporary unavailability are called {stragglers}.)
Both of these unavailabilities adversely affect service response time (latency).

A natural strategy for addressing unavailability in other domains such as communications and data storage has been through a proactive approach of adding \textit{redundancy}: making use of extra resources upfront to aid in recovery from unavailability. The effectiveness of using redundancy to reduce latency in computer systems has been shown both theoretically~\cite{joshi2014delay, gardner2015reducing, shah2016redundant, liang2013fast, wang2014efficient} as well as in practical systems~\cite{ananthanarayanan2012let, vulimiri2012more, tail-scale,rashmi2016ec}. A naive approach of adding redundancy is to replicate (that is, to have multiple copies), but this approach leads to significant resource overhead. A tool from the domain of coding theory, called \textit{erasure codes}~\cite{richardson2008modern}, provides a means for adding redundancy with significantly lesser overhead as compared to replication. Erasure codes have been successfully employed in communication~\cite{richardson2008modern}, storage~\cite{raid,facebookECsavings2010_forACM,rashmi2014hitchhiker,huang2012erasure,asterisxoring}, and distributed caching~\cite{rashmi2016ec} systems to efficiently alleviate the impact of unavailabilities.

\begin{figure}[t]
	\centering
    \includegraphics[scale=1.0]{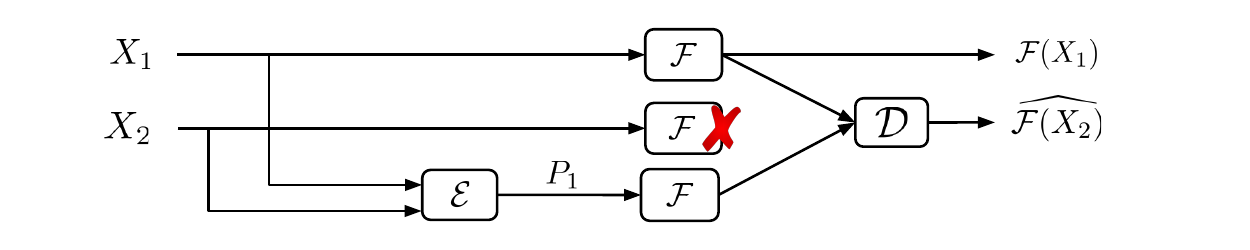}
    \caption{Coded computation with $k=2$ data inputs, and $r=1$ parity input generated by the encoding function \encodef. Here, the second output $\func(X_2)$ is unavailable (denoted by a red cross). The decoding function \decodef acts on the available outputs $\func(X_1)$ and $\func(P_1)$ to produce an approximate reconstruction of the second output denoted by $\outputReconstructTwo$.}
    \label{fig:cc}
\end{figure}

Coded computation is an emerging technique which extends the use of erasure codes to recover from \emph{unavailability of computation}. Suppose there are $k$ data inputs $\done, \dtwo,\ldots,\dk$, and suppose the goal is to apply a given function $\func$ to these $k$ data inputs, that is, to compute  $\func(\done),\func(\dtwo),\ldots,\func(\dk)$. The computations $\func(X_i)$ for different $i \in \{ 1,\ldots,k\}$ are performed on separate, unreliable devices, and hence each individual computation can straggle or fail arbitrarily.
We let $r$ represent a resilience parameter. The framework of coded computation involves two functions, an \textit{encoding function} \encodef and a \textit{decoding function} \decodef. First, the encoding function \encodef acts on the $k$ data inputs $X_1, X_2, \hdots, X_k$ to generate $r$ redundant inputs, called ``parities,'' which we denote as $P_1, P_2, \ldots,P_r$. The given function $\func$ is then applied on these $(k+r)$ inputs (data and parity) on separate, unreliable devices that can fail or straggle arbitrarily. If any $r$ or fewer outputs (out of the total $(k+r)$ outputs) are unavailable, the decoding function \decodef is applied on all the available outputs to reconstruct the unavailable ones among $\func(\done),\func(\dtwo),\ldots,\func(\dk)$. Figure~\ref{fig:cc} illustrates the coded-computation framework. Given $\func$, $k$, and $r$, the goal is to design the encoding function \encodef and the decoding function \decodef to enable reconstruction of unavailable outputs of $\func$.

Many recent works have employed erasure codes for coded computation of \textit{linear} functions such as distributed matrix-vector multiplication \linearCodedComputeRefs, and specific classes of iterative optimization algorithms~\cite{karakus2017straggler,karakus2018redundancy,tandon2017gradient}. However, to the best of our knowledge, none of the existing works are applicable for broader classes of \textit{non-linear} computations, for example, when $\func$ is a neural-network.  While the fully-connected and convolutional layers common to neural-networks are linear, they are executed along with non-linearities such as activation functions and max-pooling, effectively making the overall function non-linear. These compelling applications serve as our motivation for designing erasure codes that can handle non-linear computations.  

In the history of coding theory, advances in the design of codes have largely come about through human creativity, making use of handcrafted mathematical constructs. For coded computation of non-linear functions that arise in general tasks including machine learning applications and beyond, complex non-linear interactions make it challenging to handcraft erasure codes.
In this paper, we propose to overcome this challenge via a novel approach for designing erasure codes, that of \textit{learning codes}. 

Learning an erasure code involves learning the encoding and the decoding functions (\encodef and \decodef). Unlike the traditional approach in erasure coding, we allow the outputs of the decoding function to be an \textit{approximation} of the unavailable outputs. Approximate outputs are sufficient for many applications, such as machine learning algorithms since many of these algorithms themselves are approximate. For any input $X$ and a given function $\func$, we denote the (approximate) reconstruction  of $\outputChannel$ as $\outputReconstruct$. We make use of the ability of neural networks to perform universal function approximation by expressing the encoding and the decoding functions as neural networks. We train the neural networks for the encoding and the decoding functions in tandem via backpropagation directly through the given function $\func$.

Our approach is applicable for designing codes for imparting resilience to any differentiable non-linear  function $\func$.\footnote{Although our approach is applicable for linear functions as well, we focus primarily on non-linear functions. There are several existing works (e.g., \linearCodedComputeRefs) that address only linear functions. These approaches may be more suitable for linear functions as they guarantee exact reconstruction of unavailable outputs.} In this paper, we focus our attention on learning codes for machine learning models, specifically for the (often non-linear) computations during inference. We focus on inference as it is typically a user facing operation, and hence reducing the computation time during inference through failure and straggler mitigation has a significant impact on service quality~\cite{clipper}. In our evaluation, for the sake of concreteness, we particularly focus on functions $\func$ that are neural network models, and use the term ``\channelmodel'' to refer to $\func$. However, we emphasize that our solution extends to any differentiable function, making it applicable to a large class of tasks in machine learning and beyond.

We evaluate our framework using two neural-network based image classifiers as \channelmodel{s} (a multi-layer perceptron (MLP) and \channelResnet) using \mnist~\cite{lecun1998mnist}, \fashion~\cite{xiao2017fashion}, and \cifar~\cite{cifar} datasets. Our experimental results show that the proposed approach can \textit{accurately reconstruct a significant fraction of the unavailable outputs}: for example, 98.87\%, 92.06\%, and 80.84\% of \channelResnet classifier outputs are accurately reconstructed on \mnist, \fashion, and \cifar datasets respectively. 

Our experimental results are highly promising for the following reasons. Consider the example application of handling failures and stragglers in distributed, machine-learning inference services. Inference services typically provide strong guarantees on response times (Service Level Agreements or SLAs)~\cite{clipper}. Requests that face unavailability have prediction accuracy no better than random guessing in the absence of any corrective measures. Considering a distributed service employing \channelResnet models, if say 10\% of requests are unavailable, the overall prediction accuracy for CIFAR-10 will drop from 93.47\% to 84.12\%. Our learned codes can reconstruct the predictions for most of these unavailable cases and get close to the prediction accuracy of the underlying classifier at the cost of performing some redundant computation. Under the same scenario as considered above, learned codes can improve the overall prediction accuracy from 84.12\% to 90.59\% for \cifar, and from 89.28\% to 98.75\% for \mnist by using only 20\% redundant \channelmodel computations ($k=5, r=1$).

A note on the scope of this paper: The goal of this work is to explore the feasibility of taking a learning-based approach for designing erasure codes to impart resilience to general non-linear computations; our focus is not on optimizing the encoding and decoding function architectures for computational efficiency.
\\

The main contributions of this work are as follows:
\begin{enumerate}
	\item To the best of our knowledge, we propose the first \textit{learning-based} approach to designing erasure codes.
    \item To the best of our knowledge, we propose the first coded-computation approach for providing resilience to \textit{non-linear} functions, making it applicable to a large class of tasks in machine learning and beyond. 
    \item We carefully design neural network architectures and a training methodology for learning the encoding and decoding functions based on multilayer-perceptrons and dilated convolutional neural networks.
    \item Through extensive evaluation on two neural-network based image classifiers (a multi-layer perceptron (MLP) and \channelResnet) using  \mnist, \fashion, and \cifar datasets, we show that our learned codes can accurately reconstruct $64 - 98\%$ of the unavailable predictions.
\end{enumerate}

\section{Related Work}
A host of recent works have explored using coding theoretic approaches to impart resilience to distributed linear computations such as matrix multiplication. Lee \etal~\cite{kangwook-og} use a family of codes called  ``maximum-distance-separable'' (MDS) codes to mitigate stragglers in distributed matrix-vector multiplication. In~\cite{dutta2016short}, Dutta \etal propose Short-Dot codes to decompose long dot products that arise in certain matrix-vector multiplications into smaller products which facilitates parallel computation of such products. Li \etal~\cite{li2016unified} present a framework for navigating the tradeoff between computation time and communication time in coded computation schemes for matrix multiplication. Yu \etal~\cite{yu2017polynomial} propose Polynomial Codes for distributed matrix multiplication, which reconstruct the full matrix multiplication result using the minimal number of results from workers. Sparse Codes are introduced by Wang \etal~\cite{wang2018sparse} to exploit the sparsity of matrix operands in order to reduce decoding complexity in coded matrix-matrix multiplication. In~\cite{dutta2017coded}, Dutta \etal employ linear codes for resilient distributed convolution between two vectors. Reisizadeh \etal~\cite{reisizadeh2017coded} propose a scheme to balance the load across compute nodes for coded, distributed matrix-multiplication by taking into account heterogeneity of compute resources. In~\cite{mallick2018rateless}, Mallick \etal propose using rateless codes for distributed matrix-vector multiplication in order to make use of partial work completed by straggling nodes. In comparison to the above works which are applicable to \textit{only linear} computations, we present a learning-based approach that learns codes that can handle any differentiable \textit{non-linear} computation.

In another direction in coded computation, several recent works present approaches to using codes for providing resilience to specific iterative optimization algorithms that are employed during training of machine learning algorithms. Tandon \etal~\cite{tandon2017gradient} propose a straggler mitigation scheme for data-parallel gradient descent which involves having multiple copies of the data across the worker nodes. Under this scheme, each worker node sends a carefully constructed linear combination of its computed gradients to a master node such that the master node can complete a gradient descent iteration without having to wait for results from all the worker nodes. In~\cite{karakus2017straggler,karakus2018redundancy}, Karakus \etal propose a coded-computation approach wherein both the data and labels of a training set are encoded, and the original optimization algorithm is directly run on the encoded training dataset. For specific optimization algorithms (e.g., gradient descent and L-BFGS) and machine learning tasks (e.g., ridge regression, matrix factorization, and logistic regression), the authors present code constructions that achieve stable convergence and reduced runtime as compared to replication-based approaches. In~\cite{ldpc-matmul}, Maity \etal encode the second moment of the data matrix using LDPC codes in order to mitigate the effect of stragglers on gradient descent. The authors show that encoding the second moment reduces the number of aggregation steps necessary per training iteration compared to directly encoding the data matrix.
In contrast to these lines of work that focus on specific iterative optimization algorithms that arise during the training phase of machine learning, the focus of the our work is to add resilience through redundant computation to any differentiable non-linear computation that arise during the \textit{inference} phase of machine learning.  

Two recent works have explored taking a learning approach to designing decoding algorithms for \textit{existing} error-correcting-codes employed in the domain of communication. Nachmani \etal~\cite{dl-decoding} propose using feed-forward and recurrent neural networks for decoding a family of codes called ``block codes''. Kim \etal~\cite{rnn-decoder} show that recurrent neural networks can learn close-to-optimal decoding algorithms for several classes of well known codes employed in the domain of communication. In comparison with these works, we propose and establish the feasibility of taking a learning-based approach for the end-to-end design of codes, i.e., learning \textit{both} \textit{encoding} and \textit{decoding} algorithms. 

Another related line of work is on using neural networks for image compression and cryptography \cite{google-image-compression,tao-image-compression, crypto}.
While these lines of work are similar in spirit to learning an erasure code (transforming input data into alternate representation for later reconstruction), the overall goal, and thus the structure of the architecture and the training methodology differ significantly.

\section{Learning a Code}\label{sec:learning}

In this section, we describe our proposed approach for learning erasure codes. Recall the coded computation setup (an example of which is illustrated in Figure~\ref{fig:cc}): The encoding function \encodef acts on the $k$ data inputs to create $r$ parity inputs. The function $\func$ is then applied on these $(k+r)$ inputs (data and parity) on separate, unreliable devices that can fail or straggle arbitrarily. If any $r$ or fewer of these outputs are unavailable, the decoding function \decodef is applied on all the available outputs to reconstruct the unavailable outputs corresponding to the $k$ data inputs $\done, \dtwo, \ldots, \dk$. The goal is to learn the encoding function \encodef and the decoding function \decodef with the objective of minimizing a chosen loss function (which we discuss in more detail below in Section~\ref{sec:training}). 

We use neural networks to learn the encoding and decoding functions. We find neural networks to be a natural choice for learning the encoding and decoding functions due to their ability to perform universal function approximation~\cite{hornik1991approximation}.

In the remainder of this section,we first present our training methodology, and subsequently describe the  neural network architectures for learning the encoding and decoding functions.

\subsection{Training methodology} \label{sec:training}
Recall that our overall architecture has three functions:
the given function $\func$ whose distributed execution is to be made resilient using the learned codes, the encoding function \encodef, and the decoding function \decodef. During training the goal is to train the parameters of the neural networks for the encoding and the decoding functions. Note that the given function $\func$ is not modified during this training. 

When the given function $\func$ is a machine learning algorithm, we train the encoding and the decoding functions using the same training dataset (whenever available) that was used to train $\func$. When such a training dataset is not available, which will be the case for generic functions $\func$ outside the realm of machine learning, one can instead generate a training dataset comprising pairs $(X, \func(X))$ for various values of $X$ in the domain of $\func$. Each sample for training the encoding and decoding functions uses a set of $k$ (randomly chosen) inputs from the training dataset. 
For any sample, we perform a forward and a backward pass for each of $\binom{k+r}{r}$ possible unavailability scenarios, except for the case where all unavailable outputs correspond to parity inputs (since the only role of parities is to aid in the reconstruction of unavailable outputs corresponding to the data inputs). Any iterative optimization algorithm, such as gradient descent and its variants, may be used for training.

\begin{figure}[t]
	\centering
    \includegraphics[scale=1.2]{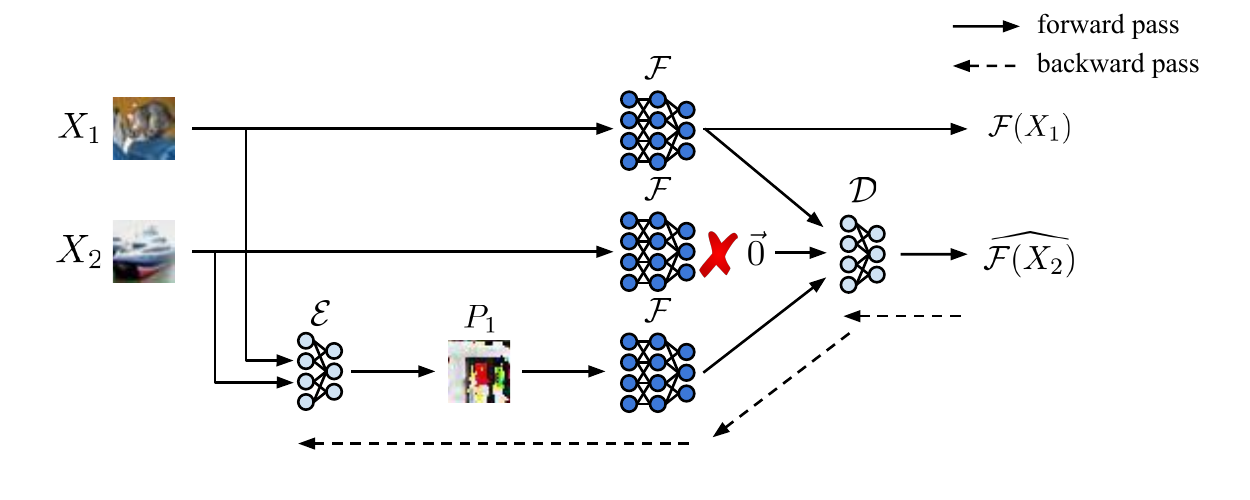}
    \caption{A forward and a backward pass in training the encoding and decoding functions for ($k=2$, $r=1$) with the given function $\func$ as a neural-network based image classifier. In the forward pass (solid left-to-right arrows), images $\done$ and $\dtwo$ are fed as inputs to the neural network for the encoding-function \encodef to generate parity image $P_1$. Each of $\done, \dtwo,$ and $P_1$ are fed through the given neural-network function $\func$. The available function outputs are fed as input to the neural network for the decoding function \decodef.  The unavailable output $\func(X_2)$ (denoted by a red cross) is replaced with a vector of zeros as input to \decodef. The decoding function \decodef produces an approximate reconstruction of $\func(X_2)$, denoted by $\widehat{\func(X_2)}$. The backward pass (dotted right-to-left arrows) propagates loss through \decodef, $\func$, and \encodef. Only the parameters of \encodef and \decodef are updated (lightly shaded neural networks).}
    \label{fig:nn-arch}
\end{figure}

A forward and a backward pass under our training method is illustrated in Figure~\ref{fig:nn-arch}.  A forward pass involves the following steps. The $k$ data inputs $\done, \dtwo, \ldots, \dk$ are fed through the encoding function to generate $r$ parity inputs $P_1, P_2, \ldots, P_r$. Each of the $(k+r)$ inputs (data and parity) are then fed through the given function $\func$. The resulting $(k+r)$ outputs $\func(X_1), \ldots, \func(X_k), \func(P_1), \ldots, \func(P_r)$  are fed through the decoding function \decodef, out of which no more than $r$ are made unavailable (discussed in detail in Section~\ref{sec:dec-unavailable}). The decoding function outputs an (approximate) reconstruction for the unavailable function outputs among $\func(X_1), \ldots, \func(X_k)$. The corresponding backward pass involves using any chosen loss function (discussed in detail below) for backpropogation through \decodef, $\func$, and \encodef. We train the encoding and decoding functions in tandem via backpropagation of losses directly through $\func$. In other words, the parameters of the encoding and the decoding functions are updated by backpropagating through $\func$. Since training backpropagates directly through $\func$, this approach is applicable to any given differentiable function $\func$. 
\\

We consider two types of losses when training the encoding and the decoding functions:
\begin{enumerate}
\item \textit{Loss with respect to function outputs:} Loss is computed between the function output $\func(X)$ and its approximate reconstruction $\widehat{\func(X)}$ produced by the decoding function. This approach can be employed for any given function $\func$.

\item \textit{Loss with respect to true labels:} When $\func$ is a machine learning algorithm, there is an additional option of calculating the loss using the true labels (when available in the training dataset). For example, consider $\func$ to be a neural network for image classification, and let $Y$ represent the true label for an input image $X$.  
Under this approach, the loss is computed between the true label $Y$ and the label predicted using $\widehat{\func(X)}$.
\end{enumerate}
The specific loss functions employed in our evaluation under both of the above approaches are discussed in Section~\ref{sec:experiments}. 

We next move on to describing the neural network architectures for learning the encoding and decoding functions.

\begin{table}
{
    \renewcommand{\arraystretch}{1.2}
\begin{subtable}[t]{0.45\textwidth}
	\centering
	\caption{\mlpcoder}
    \begin{tabular}[t]{|c|l|}
        \hline
		Layer & \multicolumn{1}{c|}{Layer Type} \\\hline
        1 &\FC $kn^2 \times kn^2$\\
        2 &\FC $kn^2 \times rn^2$\\\hline
    \end{tabular}
    \label{table:layers-mlp}
\end{subtable}
\begin{subtable}[t]{0.45\textwidth}
	\centering
    \caption{\convcoder}
    \begin{tabular}[t]{|c|l|}
    \hline
		Layer & \multicolumn{1}{c|}{Layer Type} \\\hline
 		1 & 	\Conv $3 \times 3$, \dilation 1 \\
 		2 & 	\Conv $3 \times 3$, \dilation 1 \\
		3 &		\Conv $3 \times 3$, \dilation 2 \\
		4 &		\Conv $3 \times 3$, \dilation 4 \\
		5 &		\Conv $3 \times 3$, \dilation 8 \\
		6 &		\Conv $3 \times 3$, \dilation 1 \\
 		7 &		\Conv $1 \times 1$, \dilation 1 \\\hline
    \end{tabular}
    \label{table:layers-conv}
\end{subtable}
}
    \caption{Neural network architectures for encoding functions employing fully-connected (FC) and convolutional (Conv) layers.  All convolutional layers have stride of 1. In each network, ReLU activation functions are used after all but the final layer. The activation functions are omitted above for brevity.}
    \label{table:layers}
\end{table}

\begin{figure}[!t]
	\centering
    \begin{subfigure}{0.32\textwidth}
    	\includegraphics[width=\textwidth]{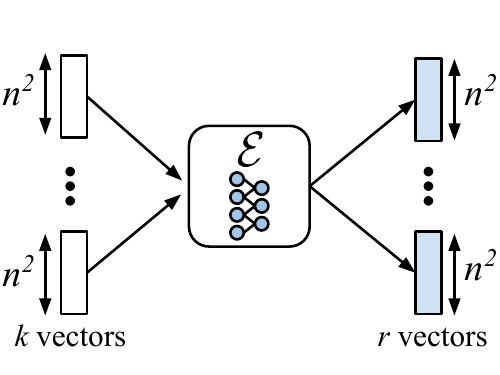}
    	\caption{\mlpcoder}
    	\label{fig:enc_mlp}
    \end{subfigure}
    \hspace{1in}
    \begin{subfigure}{0.29\textwidth}
    	\includegraphics[width=\textwidth]{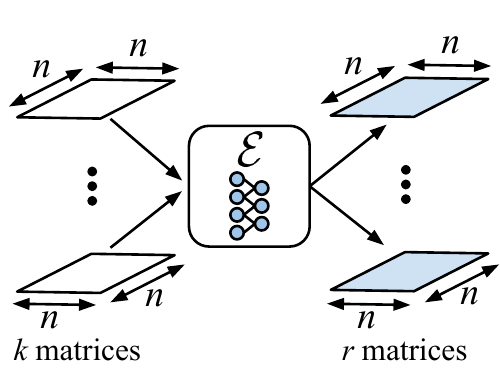}
        \vspace{0.0cm}
    	\caption{\convcoder}
    	\label{fig:enc_conv}
    \end{subfigure}
    \caption{Encoding function architecture: (a) \mlpcoder flattens inputs into $n^2$-length vectors and produces $n^2$-length parity vectors. (b) \convcoder encodes over inputs in their original dimension, as $n \times n$ matrices. The $k$ input matrices are treated as $k$ input channels.}
\end{figure}

\subsection{Encoding function architectures}\label{sec:enc}
We consider two neural network architectures for learning the encoding function 
which are applicable to any differentiable function $\func$. For concreteness, we describe the proposed architectures below by setting the given function $\func$ as a neural network for image classification over $m$ classes, and use the term ``\channelmodel'' to refer to $\func$. 
For such an $\func$, each data input $X$ is an $n \times n$ pixel image. Each function output $\func(X)$ is an $m$-length 
vector representing output from the last layer of the neural network classifier.

We now describe two neural network architectures for learning the encoding function. Recall that the encoding function acts on $k$ data inputs to create $r$ parity inputs. We first describe the architecture considering single-channel images as inputs, and consider multi-channel images in Section~\ref{sec:multi-channel}.

\subsubsection{\mlpcoder}
We first consider a simple 2-layer multilayer-perceptron (MLP) encoding function architecture, because it represents the basis for universal function approximation among neural networks \cite{hornik1991approximation}. We call this encoding function architecture \textit{\mlpcoder}. Under this architecture, the $n \times n$ data inputs are flattened into $n^2$-length vectors, as illustrated in Figure~\ref{fig:enc_mlp}. The $k$ flattened vectors from inputs $\done, \dtwo, \ldots, \dk$, are concatenated to form a single $kn^2$-length input vector to the MLP. The first fully-connected layer of the MLP produces a $kn^2$-length hidden vector. The second fully-connected layer produces an $rn^2$-length output vector, which represents the $r$ parity inputs. Each layer used in \mlpcoder is outlined in Table~\ref{table:layers-mlp}.

The fully-connected nature of the MLP allows for computation of arbitrary combinations from the $kn^2$ total inputs with a small number of layers.
While simple in design and effective for many scenarios (as will be shown in Section~\ref{sec:effects-config}), the high parameter count of the fully-connected layers can lead to overfitting. We next describe an alternate encoding function architecture that avoids overfitting, which we call \textit{\convcoder}.
 
\subsubsection{\convcoder}
The \convcoder architecture makes use of multiple convolutional layers as detailed in Table~\ref{table:layers-conv}. Unlike \mlpcoder, 
\convcoder computes over data inputs in their original $n \times n$ representation. As depicted in Figure~\ref{fig:enc_conv}, the $k$ inputs to the encoding function are treated as $k$ \textit{input channels} to the first convolution layer. This is similar to feeding the RGB representations of an image to a convolutional neural network for image classification. We explain how the encoder handles multi-channel inputs in Section~\ref{sec:multi-channel}.

\begin{figure}[!t]
	 \centering
        \begin{tabular}{@{}cc@{}}
        	\begin{subfigure}{0.15\textwidth}
            \centering
        		\includegraphics[width=\textwidth]{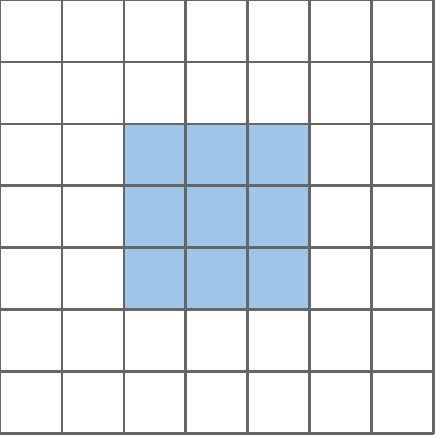}
        	\end{subfigure} 
        	&
            \begin{subfigure}{0.15\textwidth}
            \centering
        		\includegraphics[width=\textwidth]{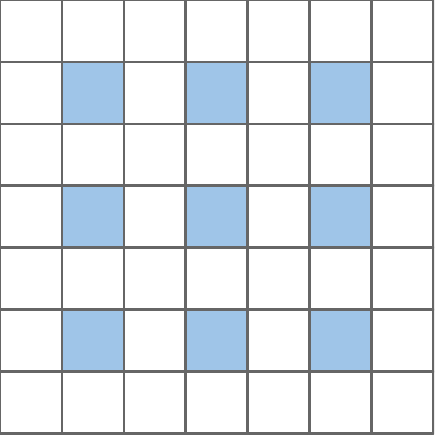}
        	\end{subfigure} \\
            dilation = 1 & dilation = 2
        \end{tabular}
        \vspace{0.1in}
        \caption{$(3\times3)$ Dilated Convolution. Traditional convolution (left) operates on adjacent cells. Dilated convolution (right) considers cells spread apart from one another.}
        \label{fig:dilation}
\end{figure}

The traditional use of convolutional layers for image classification involves repeated downsampling of an input image
to gradually expand the receptive field of convolutional filters.
This approach works well when the output dimension of the network is significantly smaller than the input dimension, which is often the case for image classification. However, the encoding function of a code produces outputs that have the same dimension as the inputs (see Figure~\ref{fig:enc_conv}). Hence, using convolutional layers with downsampling would necessitate subsequent upsampling to bring the outputs back to the input dimension. This has been shown to be inefficient in the context of image segmentation~\cite{dilated-conv}. To overcome this issue, we employ \emph{dilated convolutions} \cite{dilated-conv}. As shown in Figure~\ref{fig:dilation}, this approach increases the receptive field of a convolutional filter exponentially with linear increase in the number of layers.

Table~\ref{table:layers-conv} shows each layer of \convcoder. The first layer has $k$ input channels and the final layer has $r$ output channels, one for each parity to be produced. Each of the intermediate layers has $20k$ input channels and $20k$ output channels.
We increase the receptive field of convolutions by increasing the dilation factor, borrowing this architecture from~\cite{dilated-conv}, where it was used for image segmentation.

\convcoder uses less parameters than \mlpcoder but requires more layers to enable combinations of all input pixels. The lower parameter count compared to \mlpcoder helps avoid overfitting, as will be shown in Section~\ref{sec:effects-config}.

\subsubsection{Multi-channel input} \label{sec:multi-channel}
It is common to represent colored images as having multiple channels. For example, a $32 \times 32$ RGB image would consist of 3 channels, each $32 \times 32$ in size, representing the pixel values of each of the red, green, and blue components. Our encoding function architectures handle multi-channel inputs by encoding across each channel independently. For example, an encoding function with $k$ RGB images as inputs would encode across the $k$ red channels to produce $r$ ``red'' parity channels, and similarly for green and blue channels. The $r$ ``red'', ``green'', and ``blue'' parity channels are combined together to create $r$ parity ``RGB'' images.

\begin{figure}[!t]
	\centering
        \includegraphics[width=0.3\textwidth]{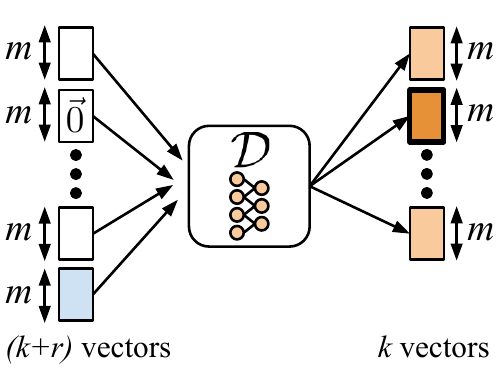}
        \caption{Decoding function architecture: The second input is unavailable and is set to a vector of zeros and the second (bolded) output represents its reconstruction. Parity inputs are shaded.}
        \label{fig:dec}
\end{figure}

\subsection{Decoding function architecture}\label{sec:dec}
As in Section~\ref{sec:enc}, for concreteness, we describe our decoding function architecture by setting the given function $\func$ as a neural network for image classification over $m$ classes. It is easy to re-purpose the proposed architecture for any differentiable function $\func$. Recall that we refer to $\func$ as the \channelmodel. The \channelmodel output $\func(X)$ for any input $X$ is an $m$-length 
vector representing output from the last layer of the neural-network classifier.
Further recall that the \channelmodel $\func$ is applied on the $(k+r)$ inputs $\done, \dtwo, \ldots, \dk, P_1,\ldots,P_r$ on separate, unreliable compute nodes that can fail or straggle arbitrarily. The decoding function \decodef operates on all the available \channelmodel outputs and reconstructs approximations of up to $r$ unavailable \channelmodel outputs among $\func(\done), \func(\dtwo), \ldots, \func(\dk)$.

Figure~\ref{fig:dec} presents the overall architecture of our decoding process. The two key design choices for the decoding function architecture are: (a) representation of the unavailable \channelmodel outputs at the input layer of the neural network for the decoding function, and (b) the neural network architecture used for learning the decoding function.

\subsubsection{Representing unavailability}\label{sec:dec-unavailable}
A key design consideration for the decoding function is in the representation of the unavailable \channelmodel outputs at its input layer. We design the decoding function to take the $(k+r$) vectors of length $m$, $\func(\done),\ldots,\func(\dk),\func(P_1),\ldots,\func(P_r)$, as inputs. Some of these inputs to the decoding function could be unavailable. In place of any unavailable input, we insert a vector of all zeros. Note that an alternative approach is to  provide the decoding function with only the (concatenated) available inputs. We chose the former as it allows us to learn a decoding function that depends on the relative position of the unavailable inputs; providing only the available inputs would hide this information. This approach is inspired by traditional (hand-crafted) erasure codes whose decoding functions leverage positional information.
Correspondingly, the output of the decoding function maintains positional information and consists of $k$ vectors
$\outputReconstructOne, \ldots, \outputReconstructK$, 
each representing an approximate reconstruction of corresponding potentially unavailable function output.

\begin{table}[t]
	\centering
    {
    \renewcommand{\arraystretch}{1.2}
    \begin{tabular}{|c|l|}
    	\hline
    	Layer & \multicolumn{1}{c|}{Layer Type} \\ \hline
        1 &\FC $(k+r)m \times km$\\
        2 &\FC $km \times km$\\
        3 &\FC $km \times km$\\ \hline 
    \end{tabular}
    }
    \caption{Neural network architecture for decoding function employing fully-connected (FC) layers. ReLU activation functions are used after all but the final layer. The activation functions are omitted above for brevity.}
    \label{table:dec-layers}
\end{table}

\subsubsection{Decoding function architecture} \label{sec:dec-layers}
We design the neural network for learning the decoding function as a 3-layer MLP as described in Table~\ref{table:dec-layers}. We use the raw outputs of the \channelmodel $\func$ as input to the decoding function. Note that we do not convert such outputs to a probability distribution (via a softmax operation) as is typically done during training of classifiers.

\section{Evaluation}\label{sec:experiments}
As discussed in Section~\ref{sec:learning}, we evaluate our approach of learning codes for imparting resilience to non-linear computations by setting the \channelmodel $\func$ as  inference on neural-network based image classifiers. For any input $X$, $\func(X)$ represents the output from the last layer of the neural network used as the \channelmodel. We start by describing our experimental setup and then present results using two neural-network based image classifiers as \channelmodel{s} (a multi-layer perceptron (MLP) and \channelResnet) on the \mnist~\cite{lecun1998mnist}, \fashion~\cite{xiao2017fashion}, and \cifar~\cite{cifar} datasets. Finally, we present a more detailed analysis of the accuracy attained by the learned codes and the quality of the predictions obtained from the reconstructed outputs.

\subsection{Experimental setup}
We implement all the encoding and decoding function architectures as well as the training methodology using PyTorch~\cite{pytorch}. Since (to the best of our knowledge) this work presents the first training methodology for learning codes for coded computation, we experiment with several loss functions and architectures, and consider multiple accuracy metrics. We describe our experimental setup below.

\subsubsection{Loss functions used in training} \label{sec:loss}
As discussed in Section~\ref{sec:training}, we use two approaches for calculating the loss when training the neural networks for the encoding and the decoding functions: (a) calculating the loss with respect to the \channelmodel output and (b) calculating the loss with respect to the true label (when available in the training dataset). When calculating the loss with respect to the \channelmodel output, we experiment with two different loss functions: (a) mean-squared error (denoted by \textit{\lossMSE}) and (b) KL-divergence (denoted by \textit{\lossKL}) between $\outputReconstruct$ and $\outputChannel$.  
When calculating loss with respect to the true labels of the underlying task, we use the cross-entropy between $\outputReconstruct$ and the true label of $X$ (denoted by \textit{\lossTrue}).

 \begin{table}[t]
	\centering
    {
    \renewcommand{\arraystretch}{1.2}
	\begin{tabular}{|c|c|c|c|}
    				\hline
      \ChannelModel & \mnist & \fashion & \cifar \\ \hline
      \channelResnet&  0.9920 & 0.9285 & 0.9347 \\
      \channelMLP	&  0.9793 &   0.8947 & -  \\ 
      \hline
    \end{tabular}
    }
    \caption{Test accuracies for the \channelmodel{s} used in our experiments on the \mnist, \fashion, and \cifar datasets.}
    \label{table:channel}
\end{table}

\subsubsection{\Channelmodel{s}}\label{sec:channelmodels}
We experiment with two neural network architectures as \channelmodel{s}: \channelMLP and \channelResnet. \channelMLP is a 3-layer multilayer-perceptron used for the \mnist and \fashion datasets containing three fully-connected layers with dimensions $784 \times 200$, $200 \times 100$, and $100 \times 10$ with ReLU activation functions following all but the final layer. We choose an MLP model due to its simplicity and its reported success on \mnist \cite{lecunn-mnist}. \channelResnet \cite{resnet} is an 18-layer state-of-the-art neural network for image classification consisting of convolutional, pooling, and fully-connected layers.\footnote{We use the  \channelResnet model described at \url{https://github.com/zalandoresearch/fashion-mnist}}. We choose to use \channelResnet for two reasons: (a) it has been shown to provide high classification accuracy on both \cifar and \fashion, and (b) it is a significantly more complex model than \channelMLP and thus provides a good alternative evaluation point for our proposed approach. Table~\ref{table:channel} shows the classification accuracies of the \channelmodel{s}. We do not use \channelMLP as a \channelmodel for \cifar as similar architectures have been shown to achieve low accuracy~\cite{cifar-mlp}.

\subsubsection{Encoding and decoding function architectures}
Recall that, in Section~\ref{sec:enc} we presented two architectures for learning the encoding function: \mlpcoder and \convcoder. Then, in Section~\ref{sec:dec}, we presented an MLP-based architecture for learning the decoding function. We present experimental results for all the proposed architectures. 

\subsubsection{Parameters and training details}
We perform experiments for all combinations of the configuration settings discussed above for $k=2$ and $k=5$ with $r=1$. We focus on $r=1$ because this corresponds to the case of typical unavailability faced in today's data centers as shown from measurements on Facebook's data analytics cluster~\cite{rashmi2013hotstorage,rashmi2014hitchhiker}. With $r=1$, the parameter settings with $k=2$ and $k=5$ correspond to 50\% and 20\% redundant computation, respectively.

Training uses minibatches of 64 samples for $k=2$ and 32 samples for $k=5$. Each sample in the minibatch consists of $k$ images from the dataset drawn randomly without replacement (i.e., no image is used more than once per epoch). Thus each minibatch for $k=2$ consists of 128 images and for $k=5$ consists of 160 images from the dataset. 
The encoding and decoding functions are trained in tandem using the Adam optimizer \cite{adam} with learning rate of 0.001 and L2-regularization of $1\times10^{-5}$. The weights for the convolutional layers are initialized via uniform Xavier initialization \cite{xavier} and weights for the fully-connected layer are initialized according to $\mathcal{N}(0, 0.01)$. All bias values are initialized to zero.

\subsubsection{Accuracy metrics}\label{sec:metric}
We measure the accuracy of the reconstructed output with respect to the machine learning task at hand using the following two metrics:
\begin{enumerate}
\item \textit{\ECacc}: This metric measures the accuracy of the reconstructed output based on its ability to recover the label predicted by the \channelmodel output. 
For example, when $\func$ is a classifier, for any input $X$, a reconstructed output $\outputReconstruct$ is considered accurate if the classes predicted using $\outputReconstruct$ and $\outputChannel$ are identical. 
More formally, let $\classificP(.)$ denote the argmax operator (which is typically used to predict the class label from the output layer of a neural network classifier). For an input $X$, a reconstructed output $\outputReconstruct$ is considered accurate if $\classificP(\outputReconstruct) = \classificP(\outputChannel)$.
This metric helps decouple the accuracy of the learned code in its ability to reconstruct unavailable \channelmodel outputs and the classification accuracy of the \channelmodel itself.

\item \textit{\Trueacc}: This metric measures the accuracy of the reconstructed output based on the true label.  
For example, when $\func$ is a classifier, for any input $X$ with true label $Y$, a reconstructed output $\outputReconstruct$ is considered accurate if the class predicted using $\outputReconstruct$ and $Y$ are identical.  
More formally, using the terminology defined above, a reconstructed output $\outputReconstruct$ is considered accurate if $\classificP(\outputReconstruct) = Y$. 
\end{enumerate}

\noindent In the results presented, for both the metrics, we calculate the aggregate accuracy by averaging the accuracy over all unavailability scenarios. If unavailability statistics are known, one can instead weigh different unavailability scenarios based on the statistics.

\subsection{Experimental results}
As discussed above, we have performed experiments for a wide range of configuration settings. To avoid clutter, we focus our discussion below on a subset of the configuration settings. The remaining experiments also have similar results as the ones discussed below and hence are relegated to appendices. 

\subsubsection{Main results} \label{sec:main-results}
We begin by discussing experiments in which training is performed using the loss with respect to the \channelmodel outputs; we focus on \ecacc as the accuracy metric. Table~\ref{table:full-results} presents the results on test datasets for all combinations of datasets, \channelmodel{s}, and parameter $k$. For clarity, we show the results only for the encoding function architecture and training loss function which achieved the highest \ecacc on the training dataset. The results for all encoding function architectures and training loss functions are available in Table~\ref{table:app-full-results} in Appendix~\ref{sec:app-full-results}.

The results in Table~\ref{table:full-results} show that our proposed approach can \textit{accurately reconstruct a significant fraction of the unavailable predictions}.  
For example, for a \channelResnet classifier on \mnist and \fashion, the learned code can accurately reconstruct 95.71\% and 82.77\% of the unavailable outputs, respectively, with only 20\% redundant \channelmodel computations (corresponding to $k=5$). Moreover, even on a more complex dataset such as \cifar, our learned code can accurately reconstruct 80.74\% of the unavailable outputs (corresponding to $k=2$).

We relegate the \trueacc attained in our experiments to the appendix. We find that the \trueacc attained by our learned codes differs only marginally from the \ecacc. The \trueacc metrics for all the experiments are available in Table~\ref{table:app-full-results} in Appendix~\ref{sec:app-full-results}. In the rest of the paper, we use the term ``accuracy'' to refer to changes in both \ecacc and \trueacc.

As mentioned earlier, our focus is on designing codes that can impart resilience for non-linear computations. There are several existing approaches (e.g., \linearCodedComputeRefs) that address  linear computations. For the sake of completeness, we include results for learning the encoding and decoding functions for a linear \channelmodel in Appendix~\ref{sec:app-logistic}. 

\subsubsection{Effect of configuration settings and parameters}~\label{sec:effects-config}
We next discuss how the accuracy attained by the learned code differs under certain parameter settings and configurations. \\

\noindent{\textbf{Value of parameter $\mathbf{k}$.} Across all datasets, \channelmodel{s}, and encoding function architectures, we find that test accuracy is significantly higher when $k=2$ than when $k=5$. We believe this is because for $k=5$, a single parity needs to pack information about 5 input images, whereas for $k=2$, a single parity contains information about only 2 inputs images. Note that with $r$ fixed, the value of $k$ controls the amount of redundant \channelmodel computation. For $r=1$, having $k=2$ corresponds to 50\% redundant \channelmodel computation and having $k=5$ corresponds to 20\% redundant \channelmodel computation. The above observation hints towards a fundamental tradeoff between \ecacc and the amount of redundant computation. The difference between $k=2$ and $k=5$ is more pronounced for the \fashion and \cifar datasets, which we attribute to the increased complexity of the dataset. \\

\noindent{\textbf{Effect of \channelmodel complexity.}} In our experiments, we find that the complexity of the \channelmodel does not have an adverse effect on the accuracy of the learned code. As discussed in Section~\ref{sec:channelmodels}, \channelResnet is a significantly more complex model  than \channelMLP, including many more layers of non-linearities. Despite this higher complexity, we see that the learned codes achieve similar accuracies for both \channelMLP and \channelResnet (in Table~\ref{table:full-results} see accuracy achieved for the two \channelmodel{s} for the \mnist and \fashion datasets). This is very promising, since it suggests that the proposed approach is effective even for complex \channelmodel{s}. 
\\

\noindent{\textbf{Encoding function architectures.}} For the \mnist and \fashion datasets, there is little difference in the accuracies attained by the two proposed neural network encoding function architectures, \mlpcoder and \convcoder. The difference between the two architectures comes to fore in the more complex dataset \cifar, where \convcoder greatly outperforms \mlpcoder. \mlpcoder's high parameter count causes it to overfit and plateau at low accuracy on \cifar, while \convcoder is able to reach significantly higher accuracy. Table~\ref{table:app-full-results} in Appendix~\ref{sec:app-full-results} contains a direct comparison between the accuracies attained by both of the encoding function architectures.\\

\begin{table}[t]
	\centering
    {
    \renewcommand{\arraystretch}{1.2}
    \begin{tabular}{|c|c|c|c|c|c|}
    	\hline
        Dataset & \ChannelModel & $k$ & \ECacc & Encoding Function Architecture & \lossShort \\ \hline
        \multirow{4}{*}{\mnist} & \multirow{2}{*}{\channelMLP}	  & 2 & 0.9885 & \mlpcoder & \lossMSE \\
        						&								  & 5 & 0.9485 & \convcoder & \lossKL \\ \cline{2-6}
                                & \multirow{2}{*}{\channelResnet} & 2 & 0.9904 & \convcoder & \lossTrue \\
                                &								  & 5 & 0.9571 & \convcoder & \lossKL \\ \hline
        \multirow{4}{*}{\fashion} & \multirow{2}{*}{\channelMLP}  & 2 & 0.9215 & \mlpcoder & \lossKL \\
        						&								  & 5 & 0.8364 & \convcoder & \lossTrue \\ \cline{2-6}
                                & \multirow{2}{*}{\channelResnet} & 2 & 0.9242 & \convcoder & \lossTrue \\
                                &								  & 5 & 0.8277 & \mlpcoder & \lossTrue \\ \hline
        \multirow{2}{*}{\cifar} & \multirow{2}{*}{\channelResnet} & 2 & 0.8074 & \convcoder & \lossMSE \\
        						&								  & 5 & 0.6431 & \convcoder & \lossMSE \\ \hline
    \end{tabular}
    }
    \caption{\ECacc (for $r=1$) on
    test data with the training loss function and the encoding function architecture chosen based on highest training \ecacc.}
    \label{table:full-results}
\end{table}

\noindent{\textbf{Loss functions used in training.}} We found only marginal difference between using the different means of calculating loss discussed in Section~\ref{sec:loss}.  Table~\ref{table:app-full-results} in Appendix~\ref{sec:app-full-results} contains results for all configurations with both loss functions.
	
\subsubsection{Detailed analysis of accuracy and quality of predictions}\label{sec:uncertainty}

We next take a deeper look at the \ecacc attained on the configurations discussed above and analyze cases where the predicted class from reconstructed outputs does not match that from the \channelmodel outputs.\\ \\

\begin{figure}[t]
	\hspace{0.125in}
    \includegraphics[width=1.0\textwidth]{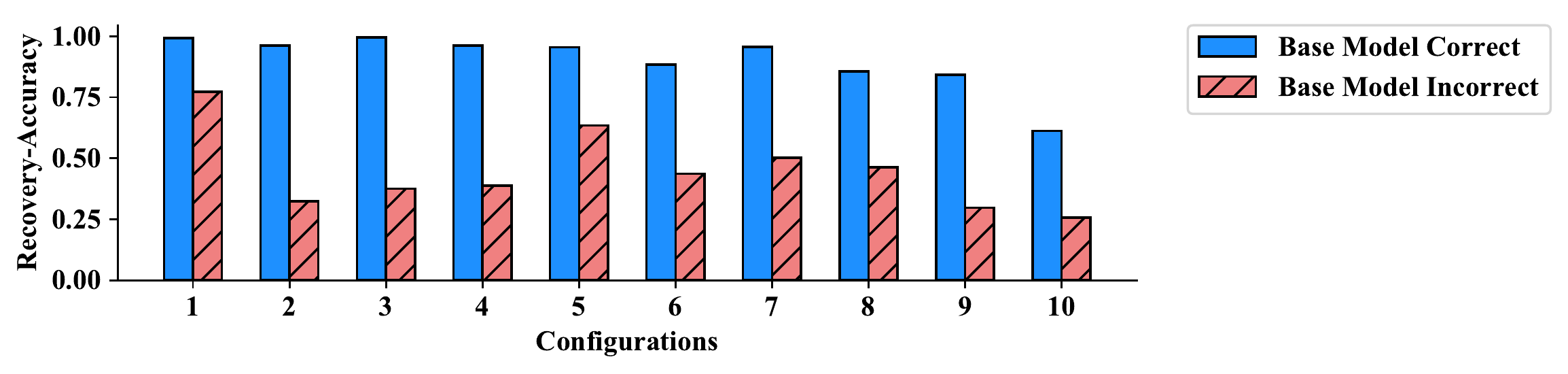}
    \captionof{figure}{\ECacc attained on samples which the \channelmodel correctly classifies (``\ChannelModel Correct'') and those which the \channelmodel incorrectly classifies (``\ChannelModel Incorrect''). The configuration for each pair of bars is available in Table~\ref{table:full-results} -- configuration number corresponds to the row number in the table. }
    \label{fig:unc}
\end{figure}

\begin{figure}[t]
	\hspace{0.125in}
    \includegraphics[width=1.0\textwidth]{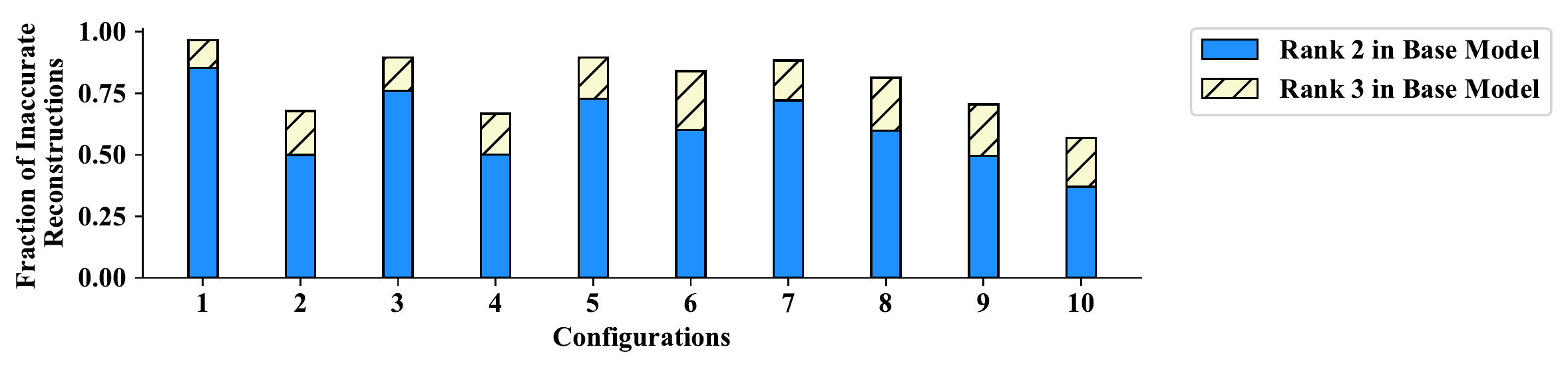}
    \caption{Fraction of inaccurate reconstructions that are rank 2 and 3 in the \channelmodel output. The configuration for each bar is available in Table~\ref{table:full-results} -- configuration number corresponds to the row number in the table.} 
    \label{fig:unc_rank}
\end{figure}

\noindent{\textbf{\ECacc stratified based on accuracy of the \channelmodel.} In our experiments, interestingly, the learned codes achieve a significantly higher \ecacc on the set of samples that the \channelmodel classifies correctly as compared to the set of samples that the \channelmodel classifies incorrectly. Figure~\ref{fig:unc} shows the \ecacc on these two sets of samples in all the configurations of experiments listed in Table~\ref{table:full-results}. We see that the learned codes achieve, on average, $2.19$ times higher \ecacc on the set of samples that the \channelmodel classifies correctly (``\ChannelModel Correct'' in Figure~\ref{fig:unc}) as compared to the set of samples that the \channelmodel classifies incorrectly (``\ChannelModel Incorrect'' in Figure~\ref{fig:unc}). Thus, the \ecacc of the learned codes is higher on samples where it indeed matters more to reconstruct accurately. \\

\noindent{\textbf{Analysis of errors in the learned code.}}
Here we analyze how poor is the class predicted from  inaccurate reconstructions. Specifically, we look at the samples for which the reconstructed output is inaccurate with respect to the \channelmodel output (as defined under \ecacc in Section~\ref{sec:metric}), and analyze how far the resulting predicted class label is from the label predicted by the \channelmodel output. We quantify the quality of the label predicted from an inaccurate reconstruction by its rank in the \channelmodel output. A rank of 2 means that the class predicted using the reconstruction was ranked second in the \channelmodel output.\footnote{Note that rank 1 is unattainable since we are analyzing only those instances for which the predicted class from the reconstruction does not match that of the \channelmodel output.} Figure~\ref{fig:unc_rank} shows the fraction of inaccurate reconstructions which lead to predicted labels that have rank 2 and rank 3 in the \channelmodel output for the configurations considered in Table~\ref{table:full-results}. We see that, on average, $61.29\%$ of the inaccurate reconstructions result in a class prediction that is the second best in the \channelmodel output. Furthermore, on average, $79.12\%$ of the inaccurate reconstructions result in a class prediction among the top 3 predictions of the \channelmodel output. Thus, even when the class prediction resulting from a reconstructed output does not match that of the \channelmodel output, the predicted class is not far off.

\section{Conclusion}
Coded computation is an emerging technique which makes use of coding-theoretic tools to impart resilience against failures and stragglers in distributed computation. However, the applicability of current techniques to general computation,  including machine learning algorithms, is limited due to the lack of codes that can handle non-linear functions. In this paper, we propose a novel learning-based approach for designing erasure codes that approximate unavailable outputs for any differentiable non-linear function. We present carefully designed neural network architectures and a training methodology for learning the encoding and decoding functions. We show that our learned codes can accurately reconstruct up to 98.85\%, 92.15\%, and 80.74\% of the unavailable class predictions from image classifiers for \mnist, \fashion, and \cifar datasets, respectively. These results are highly promising as they show the potential of using learning-based approaches for designing erasure codes and herald a new direction for coded computation by handling general non-linear computations.

\bibliographystyle{alpha}
\bibliography{references}

\newpage
\appendix
\begin{appendices}

\newpage
\section{Full Experimental Results}\label{sec:app-full-results}
Table~\ref{table:app-full-results} contains experimental results for all configurations considered in Section~\ref{sec:experiments}.

\begin{table}[h!]
	\centering
    {
    \renewcommand{\arraystretch}{1.2}
    \begin{tabular}{|c|c|c|c|cc|cc|}
    	\hline
        \multicolumn{4}{|c|}{} & \multicolumn{2}{c|}{\mlpcoder} &  \multicolumn{2}{c|}{\convcoder} \\ \hline
        \multirow{2}{*}{Dataset} & \multirow{2}{*}{\ChannelModel} & \multirow{2}{*}{$k$} & \multicolumn{1}{p{2.5cm}|}{\centering Training Loss \\ Function} & \multicolumn{1}{p{1.5cm}}{\centering Recovery \\ Accuracy} & \multicolumn{1}{p{1.5cm}}{\centering Overall \\ Accuracy} & \multicolumn{1}{|p{1.5cm}}{\centering Recovery \\ Accuracy} & \multicolumn{1}{p{1.5cm}|}{\centering Overall \\ Accuracy} \\ \hline
        \multirow{12}{*}{\mnist} & 	\multirow{6}{*}{\channelMLP}	& \multirow{3}{*}{2} & \lossKL & 0.9769 	& 0.9758 & 0.9831	& 0.9854 \\
                                &									& 					 & \lossMSE & 0.9885 	& 0.9776 & 0.9767 	& 0.9770 \\
                              	&									&					 & \lossTrue& 0.9737	& 0.9768 & 0.9769	& 0.9893 \\ \cline{3-8}
                                &									& \multirow{3}{*}{5} & \lossKL & 0.9371 	& 0.9340 & 0.9485 	& 0.9518 \\
                                &									& 					 & \lossMSE & 0.9480 	& 0.9424 & 0.9339 	& 0.9357 \\
                              	&									&					 & \lossTrue&	0.9251  & 0.9232 & 0.9474	& 0.9533 \\ \cline{2-8}
                                & 	\multirow{6}{*}{\channelResnet}	& \multirow{3}{*}{2} & \lossKL & 0.9742 	& 0.9760 & 0.9836 	& 0.9854 \\
                                &									& 					 & \lossMSE & 0.9788 	& 0.9806 & 0.9887 	& 0.9888 \\
                              	&									&					 & \lossTrue&	0.9774	& 0.9796 & 0.9904   & 0.9925 \\ \cline{3-8}
                                &									& \multirow{3}{*}{5} & \lossKL & 0.9460 	& 0.9466 & 0.9571 	& 0.9585 \\
                                &									& 					 & \lossMSE & 0.9349 	& 0.9359 & 0.9415 	& 0.9433 \\
                              	&									&					 & \lossTrue&	0.9401  & 0.9407 & 0.9171   & 0.9178 \\ \hline
        \multirow{12}{*}{\fashion}& 	\multirow{6}{*}{\channelMLP}	& \multirow{3}{*}{2} & \lossKL &0.9215& 0.8800 & 0.9128 	& 0.9080 \\
                                &									& 					 & \lossMSE & 0.8484 	& 0.8196 & 0.8471 	& 0.8253 \\
                              	&									&					 & \lossTrue&	0.9107	& 0.8808 & 0.9036	& 0.9185 \\ \cline{3-8}
                                &									& \multirow{3}{*}{5} & \lossKL & 0.8275 	& 0.7997 & 0.8300 	& 0.8153 \\
                                &									& 					 & \lossMSE & 0.7133 	& 0.6987 & 0.7302 	& 0.7193 \\
                              	&									&					 & \lossTrue&	0.8259  & 0.8037 & 0.8364   & 0.8282 \\ \cline{2-8}
                                & 	\multirow{6}{*}{\channelResnet}	& \multirow{3}{*}{2} & \lossKL & 0.9002 	& 0.8845 & 0.9206 	& 0.9031 \\
                                &									& 					 & \lossMSE & 0.8960 	& 0.8815 & 0.8982 	& 0.8892 \\
                              	&									&					 & \lossTrue&	0.8947  & 0.8880 & 0.9242   & 0.9164 \\ \cline{3-8}
                                &									& \multirow{3}{*}{5} & \lossKL& 0.8219 	& 0.8133 & 0.8033 	& 0.7960 \\
                                &									& 					 & \lossMSE & 0.7726 	& 0.7672 & 0.7939 	& 0.7885 \\
                              	&									&					 & \lossTrue&	0.8277  & 0.8203 & 0.8303   & 0.8248 \\ \hline
       \multirow{6}{*}{\cifar}  & 	\multirow{6}{*}{\channelResnet}	& \multirow{3}{*}{2} & \lossKL & 0.4293 	& 0.4283 & 0.7889 	& 0.8002 \\
                                &									& 					 & \lossMSE & 0.4107 	& 0.4116 & 0.8074 	& 0.8204 \\
                              	&									&					 & \lossTrue&	0.4284  & 0.4238 & 0.7980   & 0.8106 \\ \cline{3-8}
                                &									& \multirow{3}{*}{5} & \lossKL & 0.1889 	& 0.1895 & 0.5368 	& 0.5382 \\
                                &									& 					 & \lossMSE & 0.1913 	& 0.1936 & 0.6431 	& 0.6466 \\
                              	&									&					 & \lossTrue&	0.1874  & 0.1890 & 0.5224   & 0.5287 \\
                                \hline
    \end{tabular}
    }
    \caption{\ECacc and \trueacc for $r=1$ for all configuration settings.}
    \label{table:app-full-results}
\end{table}

Recall that results presented in Section~\ref{sec:main-results} did not consider all parameter settings and configurations. We briefly highlight some relevant configuration comparisons made available through the full results presented in Table~\ref{table:app-full-results}. \\

\noindent{\textbf{\Trueacc metric:}} Looking at the ``Recovery Accuracy'' and ``Overall Accuracy'' columns of Table~\ref{table:app-full-results}, there is little difference between the two metrics, when holding architecture, parameters, and other configuration settings constant. We believe that the similarity of these two metrics can in part be explained by the observation in Section~\ref{sec:uncertainty} that the \ecacc attained on samples which are correctly classified by the \channelmodel is often significantly higher than that attained on samples which are incorrectly classified by the \channelmodel. \\
    
\noindent\textbf{Difference between training loss functions.} 
Results with ``\lossTrue'' as the training loss function in Table~\ref{table:app-full-results} correspond to those configurations for which training calculated loss via cross-entropy between the reconstructed output $\outputReconstruct$ and the true label of $X$. The \ecacc and \trueacc for the \lossTrue configurations are very similar to those of the corresponding configurations with \lossKL and \lossMSE (which calculate the KL-divergence and MSE, respectively, between $\outputReconstruct$ and $\outputChannel$).

There are two configurations for which we observe significant difference in the accuracies attained using each loss function. First, for \fashion with \channelMLP as the base model and \mlpcoder as the encoding function architecture, we find that using \lossMSE leads to a decrease in test \ecacc and \trueacc compared to both \lossKL and \lossTrue. 
Second, when training \channelResnet \channelmodel{s} on \cifar with \convcoder for $k=5$, we find that using \lossMSE leads to roughly 0.10 increase in test \ecacc and \trueacc compared to using \lossKL and \lossTrue.

\section{Multinomial Logistic Regression} \label{sec:app-logistic}
In this section, we evaluate our learned codes on a multinomial logistic regression problem on the \mnist dataset. The overall calculation in multinomial logistic regression is of the form $\sigma(AX + b)$ for parameters $A$ and $b$, data $X$, and with $\sigma$ being a softmax operator.  Recall from Section~\ref{sec:dec-layers} that the inputs to our neural network decoding function are the raw outputs of the \channelmodel (prior to any softmax operation), which are not converted to a probability distribution. As such, the available inputs to the decoding function are $\outputChannel = (AX + b)$. The softmax operation is applied to reconstructed outputs of the decoder. For the \mnist dataset, the \channelmodel thus consists of parameter matrix $A \in \mathbb{R}^{10 \times 784}$ and a vector $b \in \mathbb{R}^{10}$. The value 10 corresponds to the number of classes in the \mnist dataset. Each 28 $\times$ 28 input image from the \mnist dataset is flattened to form a 784 length vector $X$. 

We train the \channelmodel described above on the \mnist dataset. The trained \channelmodel achieves an accuracy of 0.9283 on the \mnist test set.
\begin{table}[t]
	\centering
    {
    \renewcommand{\arraystretch}{1.2}
	\begin{tabular}{|c|c|c|c|}
    \hline
    Encoding Function Architecture 	& k & \ECacc & \Trueacc\\ \hline
    \multirow{2}{*}{\mlpcoder}		& 2 & 0.9831 & 0.9279 \\
    								& 5 & 0.9817 & 0.9270 \\ \hline
    \multirow{2}{*}{\convcoder}		& 2 & 0.9899 & 0.9295 \\
    								& 5 & 0.9869 & 0.9260 \\ \hline   
    \end{tabular}
    }
    \caption{Test \ecacc and \trueacc for logistic regression \channelmodel on the \mnist dataset with $r=1$ and using KL-divergence as the loss function.}
    \label{table:app-logistic}
\end{table}
Table~\ref{table:app-logistic} shows the \ecaccs achieved on the test set over this \channelmodel by each of the encoding function architectures described in Section~\ref{sec:enc} with $r=1$, $k$ being 2 and 5, and using KL-divergence as the loss function. In all cases, the proposed codes are able to achieve a high \ecacc. We note that $\func(X)=(AX+b)$ can be (trivially) transformed into a linear function by juxtaposing the matrix A and the vector b and appending 1 to vector X. Hence even though our approach provides high \ecacc, the existing approaches that address only linear functions~\linearCodedComputeRefs are perhaps more suitable for this particular $\func$ as these approaches guarantee exact reconstruction of unavailable outputs. However, note that these existing approaches are applicable only for linear functions while our goal is to handle non-linear functions.

\end{appendices}
\end{document}